%% file: main_neurips_2022.tex
\documentclass[width=15cm]{article}
\usepackage[a4paper, total={6in, 10 in}]
{geometry}
\input{preamble}

\usepackage{subcaption}
\usepackage{appendix}
\usepackage{multirow}

\usepackage[utf8]{inputenc} 
\usepackage[T1]{fontenc}    
\usepackage{hyperref}       
\usepackage{url}            
\usepackage{booktabs}       
\usepackage{amsfonts}       
\usepackage{nicefrac}       
\usepackage{microtype}      
\usepackage{xcolor}         

\title{Unraveling overoptimism and publication bias in ML-driven science}

\author{
Pouria Saidi$^{a}$\thanks{e-mail:psaidi@asu.edu},
Gautam Dasarathy$^{a}$,
Visar Berisha$^{a,b}$ \and
$^{a}$School of Electrical, Computer and Energy Engineering,\\ Arizona State University Tempe, USA, 85281\\  $^{b}$College of Health Solutions, Arizona State University, Tempe, USA, 85281}

\date{}
\begin{document}
\maketitle
\begin{abstract}
Machine Learning (ML) is increasingly used across many disciplines with impressive reported results. However, recent studies suggest published performance of ML models are often overoptimistic. Validity concerns are underscored by findings of an inverse relationship between sample size and reported accuracy in published ML models, contrasting with the theory of learning curves where accuracy should improve or remain stable with increasing sample size. 
This paper investigates factors contributing to overoptimism in ML-driven science, focusing on overfitting and publication bias. We introduce a novel stochastic model for observed accuracy, integrating parametric learning curves and the aforementioned biases. We construct an estimator that corrects for these biases in observed data. Theoretical and empirical results show that our framework can estimate the underlying learning curve, providing realistic performance assessments from published results. Applying the model to meta-analyses of classifications of neurological conditions, we estimate the inherent limits of ML-based prediction in each domain.
\end{abstract}

\input{TEX_files/Introduction}
\input{TEX_files/Results}

\input{TEX_files/MaterialsandMethods}
\input{TEX_files/conclusion}
\bibliographystyle{IEEEbib}
\bibliography{bibfile}
\newpage
\appendix
\input{TEX_files/Appendix1}

\end{document}

%% file: preamble.tex
\usepackage[mathscr]{eucal}
\usepackage[cmex10]{amsmath}
\usepackage{epsfig,epsf,psfrag}
\usepackage{amssymb,amsmath,amsfonts,latexsym}
\usepackage{amsmath,graphicx,bm,xcolor,url}
\usepackage{epstopdf}
\usepackage{array}
\usepackage{verbatim}
\usepackage{bm}
\usepackage{algorithmic, cite}
\usepackage{algorithm}
\usepackage{verbatim}
\usepackage{textcomp}
\usepackage{mathrsfs}
\usepackage{amsthm}
\usepackage{cancel}

\catcode`~=11 \def\UrlSpecials{\do\~{\kern -.15em\lower .7ex\hbox{~}\kern .04em}} \catcode`~=13 

\allowdisplaybreaks[3]

\newcommand{\bw}{\mathbf{w}}

\newcommand{\bx}{\mathbf{x}}

\newcommand{\by}{\mathbf{y}}

\newcommand{\bz}{\mathbf{z}}

\DeclareMathAlphabet{\mathbsf}{OT1}{cmss}{bx}{n}
\DeclareMathAlphabet{\mathssf}{OT1}{cmss}{m}{sl}

\DeclareSymbolFont{bsfletters}{OT1}{cmss}{bx}{n}  
\DeclareSymbolFont{ssfletters}{OT1}{cmss}{m}{n}
\DeclareMathSymbol{\bsfGamma}{0}{bsfletters}{'000}
\DeclareMathSymbol{\ssfGamma}{0}{ssfletters}{'000}
\DeclareMathSymbol{\bsfDelta}{0}{bsfletters}{'001}
\DeclareMathSymbol{\ssfDelta}{0}{ssfletters}{'001}
\DeclareMathSymbol{\bsfTheta}{0}{bsfletters}{'002}
\DeclareMathSymbol{\ssfTheta}{0}{ssfletters}{'002}
\DeclareMathSymbol{\bsfLambda}{0}{bsfletters}{'003}
\DeclareMathSymbol{\ssfLambda}{0}{ssfletters}{'003}
\DeclareMathSymbol{\bsfXi}{0}{bsfletters}{'004}
\DeclareMathSymbol{\ssfXi}{0}{ssfletters}{'004}
\DeclareMathSymbol{\bsfPi}{0}{bsfletters}{'005}
\DeclareMathSymbol{\ssfPi}{0}{ssfletters}{'005}
\DeclareMathSymbol{\bsfSigma}{0}{bsfletters}{'006}
\DeclareMathSymbol{\ssfSigma}{0}{ssfletters}{'006}
\DeclareMathSymbol{\bsfUpsilon}{0}{bsfletters}{'007}
\DeclareMathSymbol{\ssfUpsilon}{0}{ssfletters}{'007}
\DeclareMathSymbol{\bsfPhi}{0}{bsfletters}{'010}
\DeclareMathSymbol{\ssfPhi}{0}{ssfletters}{'010}
\DeclareMathSymbol{\bsfPsi}{0}{bsfletters}{'011}
\DeclareMathSymbol{\ssfPsi}{0}{ssfletters}{'011}
\DeclareMathSymbol{\bsfOmega}{0}{bsfletters}{'012}
\DeclareMathSymbol{\ssfOmega}{0}{ssfletters}{'012}

\newtheorem{theorem}{Theorem} 
\newtheorem{lemma}[theorem]{Lemma}

\newtheorem{data model}{Data Model}

\newcommand{\qednew}{\nobreak \ifvmode \relax \else
      \ifdim\lastskip<1.5em \hskip-\lastskip
      \hskip1.5em plus0em minus0.5em \fi \nobreak
      \vrule height0.75em width0.5em depth0.25em\fi}

%% file: TEX_files/Introduction.tex
\section{Introduction} \label{sec:Intro}
Recent advancements in machine learning (ML) have opened new avenues for research across many disciplines, giving rise to the field of ML-based science (e.g. sociology \cite{rosenbusch2021supervised}, medicine \cite{vandewiele2021overly}, education \cite{alenezi2020utilizing} and digital health \cite{mathews2019digital}). The rapid adoption of ML in these fields is driven in large part by high reported accuracies in academic publications. 

\begin{figure*}
    \centering
    \begin{subfigure}[b]{0.32\textwidth}
    \includegraphics[width = \textwidth]{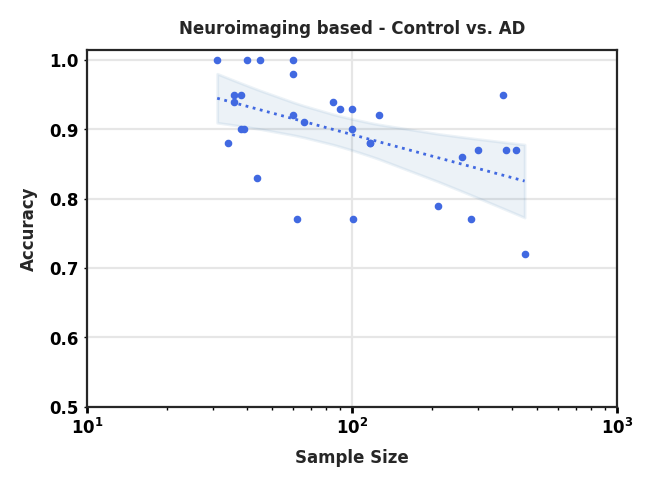}
    \caption{}
    \label{subfig:loglin_AD}
    \end{subfigure}
    \hfill
    \begin{subfigure}[b]{0.32\textwidth}
    \includegraphics[width = \textwidth]{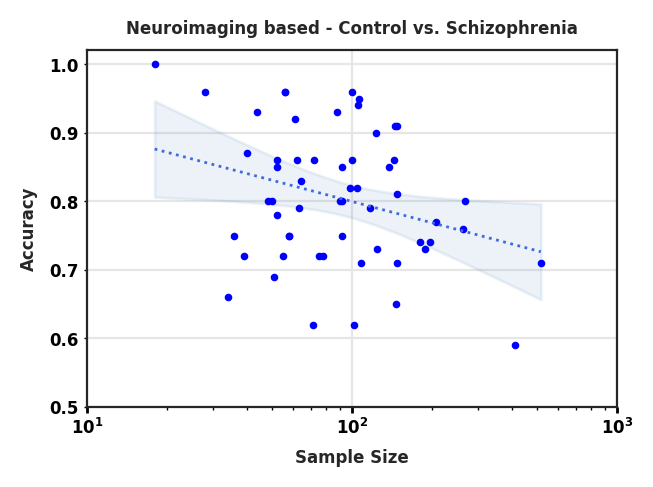}
    \caption{}
    \label{subfig:loglin_SCH}
    \end{subfigure}
    \hfill
    \begin{subfigure}[b]{0.32\textwidth}
    \centering
    \includegraphics[width = \textwidth]{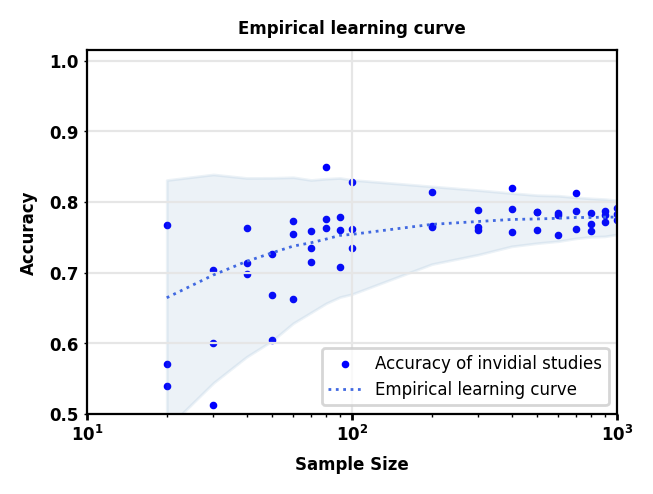}
    \caption{}
    \label{fig:TLC}
\end{subfigure}
    \caption{The reported accuracy vs. sample size from a collection of results in a meta-analysis study \cite{arbabshirani2017single}. This analysis considers neuroimaging-based classification models between a control group and a patient cohort with (a) Alzheimer's disease (AD), and (b) Schizophrenia. (c) An empirical learning curve that is obtained by solving a binary classification problem using different ML models and sample sizes. The y-axis is in linear
scale and the x-axis is in log scale.}
    \label{fig:negative_asc}
\end{figure*}

Despite impressive reported results, several recent studies have raised questions about their validity \cite{berisha2022over,arbabshirani2017single,kapoor2023leakage}. For instance, a collection of results from a survey of prediction of brain disorders \cite{arbabshirani2017single} reveal an unexpected negative association between sample size and reported accuracy in these studies. In Fig. \ref{subfig:loglin_AD} and \ref{subfig:loglin_SCH} we illustrate this negative relationship for ML-based studies focused on prediction of Alzheimer's Disease (AD) and Schizophrenia. A similar trend has been reported in \cite{berisha2021digital}, where the authors analyzed published accuracies of speech-based ML models to predict AD and other forms of cognitive impairment (CI) \cite{de2020artificial,petti2020systematic,martinez2021ten}, and in \cite{vabalas2019machine}, where the authors analyze the performance of ML models for detection of Autism spectrum disorder (ASD). This is in contrast with our theoretical understanding of machine learning, where increasing sample sizes should not decrease the accuracy of a properly trained model \cite{berisha2021digital,viering2022shape}. Fig. \ref{fig:TLC} illustrates a well-behaved learning curve that follows this intuition, and is obtained by solving a binary classification problem using properly-trained and evaluated ML models with randomly sampled datasets of different sample size.

The authors in \cite{berisha2021digital} postulate that overfitting and publication bias give rise to the negative association between reported accuracy and sample size. Overfitting occurs when a model captures not only the underlying patterns in the training data but also its noise and idiosyncrasies, leading to poor generalization on new, unseen data. This issue is especially pronounced in situations with limited data. It can become challenging to identify overfitting to the test set. This can happen due to unintended interdependencies between training and test datasets that can arise during model development \cite{kaufman2012leakage,kapoor2023leakage}. Model development is inherently an iterative, adaptive process \cite{dwork2015preserving} where researchers reuse tests sets to refine models. However, repeated use of the same test set can lead to inadvertently learning specific patterns or noise unique to that set, exacerbating overfitting issues. This problem has gained considerable attention recently in ML-based science \cite{kapoor2023leakage}, with the recognition that it likely results in overestimation of  model performance and unrealistic reported accuracy, particularly at small sample sizes \cite{roelofs2019meta}.

A less-studied cause of overoptimism in ML-based papers is publication bias. It is known that training an ML model with limited data typically results in high variance accuracy estimates \cite{james2013introduction}. This increased variability can lead to instances where the accuracy of the model is both underestimated and overestimated. However, models with higher estimated accuracy are more likely to be published \cite{serra2021nonreplicable}, a phenomenon known as publication bias or the file drawer effect in the social and medical sciences \cite{rosenthal1979file}. As a result, models with inflated estimates of the performance are more likely to be published. Meanwhile, models trained with larger sample sizes produce results that are more accurate and exhibit less variability, making them a more reliable measure of accuracy. Therefore, both overfitting and publication bias have a greater impact on models when the sample size is small, and this effect diminishes as the sample size increases. Hence, we posit that the negative association observed in Fig. \ref{fig:negative_asc} is attributable to overfitting and publication bias. Both causes of overoptimism are likely compounded by the incentives created by the academic community’s outsized emphasis on high accuracy as a primary reason for publication of new methodologies in ML-based science and the increased analytical flexibility of ML methods \cite{ioannidis2005most}. 

Overoptimistic reported accuracy results create challenges for true scientific progress and responsible deployment of ML models. They create a skewed perception of the state of knowledge in the field and inflate expectations for the practical application of research. These inflated expectations can lead to sensationalized stories in the press and premature deployment  \cite{raji2022fallacy,yawer2023reliability}. When these models fail after deployment, these expectations are not met and can negatively impact the public's trust in this technology \cite{jobin2019global}. We posit that this problem may be amplified for some fields with the new federal public access policy mandates access to the results of federally funded research \cite{mem2013,mem2022}. Ioannidis  \cite{ioannidis2005most} discussed that rapidly evolving scientific fields with more scientific teams involved and with greater flexibility in design and analytical methods, have higher chances to report false and overoptimistic research findings. Availability of new data in ML-based science will likely attract more scientific teams and consequently, this may lead to an increase in the number of publications. 
This has the potential to set the stage for a perfect storm where deciding whether the new results are trustworthy or not becomes more difficult, even for the experts in the field.

In this paper, we present a novel observation model for the \textit{published} classification accuracy of ML models based on the notion of parametric learning curves, taking into account both overfitting and publication bias. We leverage this model and further propose a solution to alleviate the overoptimism and determine realistic estimates of model performance by correcting the bias due to both causes. 

\subsection{Proposed model} 
\label{subsec:proposed_model} 
We introduce the main idea of our modeling approach here. A more detailed explanation appears in the Methods section. The relationship between training dataset size $n$ and a machine learning model's performance $y(n)$ (say classification accuracy) is known to be effectively modeled using equations of the form $y(n) = A + \alpha n^{\beta}$ \cite{hoiem2021learning,cortes1993learning,john1996static,hestness2017deep,kolachina2012prediction}. Such models, known as \emph{power laws}, have found application in a variety of areas such as network science \cite{artico2020PLM3}, social science \cite{SocialScience2018PLM2} and infection disease surveillance \cite{Meyer2014PLM1}.

Consider a scenario where a research team has access to a dataset of sample size $n$ for the development of a binary classification model. During ML model development, the team iterates over the dataset multiple times, adjusting parameters and algorithms based on insights gained from previous runs, resulting in the final estimate of model accuracy. We propose an observation model for such estimates of the accuracy as
\begin{equation}
\label{eq:parametric_model}
    \by^{\star}_n = A + \alpha n^{\beta} + \bw_n \:,
\end{equation}
 where $\by^{\star}_n$ is the classification accuracy of an ML model given the sample size - $n$,
\begin{itemize}
    \item $A$ is the the limiting performance
    \item $\beta<0$ is the learning rate
   \item $\alpha<0$ is the power law index
   \item $\bw_n \sim \mathcal{N}\left(\mu_n,\sigma_n^2\right)$ is a random variable with a Gaussian distribution. The idea behind adding $\bw_n$ is twofold. The random variable $\bw_n$ accounts for the inherent variations in the estimates of the model accuracy and the effects of overfitting on estimated accuracy. Variations in model accuracy arise due to differences in data selection and data splitting for model training and testing, and they  scale with $1/\sqrt{n}$ \cite{arora2021rip,vapnik2006estimation}. We incorporate this information and use $\sigma_n = c_1/\sqrt{n}$, where $c_1$ is a constant, to quantitatively model the variations in model accuracy. Similarly, previous studies have shown that overfitting to test set inflates estimated model accuracies with $n^{-0.5}$. As a result, we adjust the mean of the Gaussian distribution to $\mu_n = \zeta n^{-0.5}$ \cite{dwork2015preserving,arora2021rip} to model this. Here, $\zeta$ denotes the overfitting index, or the bias in the average accuracy estimate due to overfitting.
\end{itemize}
 
The model proposed in Eq. \eqref{eq:parametric_model} posits that the average performance of the model can be represented as $A + \alpha n ^{\beta}$ in the absence of overfitting. By applying this model across a range of sample sizes, the parametric model uncovers the true estimate of the learning trajectory, or learning curve, of the ML model.

A research team may elect to publish the observed results or not, depending on whether they deem the resulting accuracy sufficiently high for publication. Such self-selection has been shown to cause a bias in the published literature \cite{easterbrook1991publication}. We model publication bias in the \textit{reported} classification accuracies using a \emph{selection mechanism} \cite{amemiya1984tobit}. Under this model, a research team decides to publish their result if the classification accuracy $\by^{\star}_n > \gamma_n$, where $\gamma_n$ is a threshold that depends on the sample size. This operation models, for instance, self-selection by the authors and peer review. We posit that this process is carried out independently by multiple research teams at various levels of data availability.  The selection model, therefore, implies that consumers of this literature only observe a biased sample of accuracies that are greater than some threshold.

In this work, we aim to estimate the parameters of the parametric learning curve in Eq. \eqref{eq:parametric_model} from the observed classification accuracies. Due to the censoring of observations from publication bias, conventional methods such as ordinary least squares lead to the unrealistic negative association observed in Fig. \ref{fig:negative_asc}. Therefore, we propose a new solution based on truncated regression to estimate the parameter values and provide theoretical and empirical results that demonstrate we can reliably estimate the true learning curve from a series of overoptimistic observations. 

\subsection{Contributions} The following summarizes the main contributions of this paper:
\begin{itemize}
\item We propose a novel observation model for the \textit{published} classification accuracy of ML models. This model is based on the notion of parametric learning curves that can be represented using power law models, and it accounts for overfitting and publication bias as two influencing factors for overoptimism in the ML literature.
    \item We propose a solution based on truncated regression and show theoretically that it is possible to identify the learning trajectory of ML models even without prior information about the selection model. Furthermore, we use the observation model to devise a cost function for estimating the true learning curve from overoptimistic accuracy results.
    \item We apply the model to different meta-analyses in the digital health literature. Particularly, we consider meta-studies of brain disorders, including Alzheimer's disease (AD) and other forms of cognitive impairment (CI), Schizophrenia, attention deficit hyperactivity disorder (ADHD), and Autism spectrum disorder (ASD); the reported ML model results are based on multiple modalities such as 
    neuroimaging and speech data. Our analysis highlights the prevalence of overoptimism in these fields and provides realistic estimates of ML model performance in each field.
   
\end{itemize}

%% file: TEX_files/Results.tex
\section{Results}
\label{sec:results}

We evaluate the observation model and the accuracy of the solution in estimating ML model performance under overoptimism. The evaluation comprises three interconnected experiments, each highlighting a key component of the model's performance and relevance.

 Experiment 1 serves as a foundational test, where we generate data directly from the observation model and estimate the known underlying learning curve. This experiment establishes the basic efficacy of our model, demonstrating its capability to accurately recover learning curve parameters in a controlled environment. This sets the stage for more complex scenarios in subsequent experiments.

Experiment 2 is a realistic simulation of the real-world process of ML model development, integrating elements of overfitting and selective reporting based on a minimal accuracy threshold. We consider binary classification problems and first empirically derive the true learning curve by progressively increasing sample sizes in ML model training. Next we model overfitting to the test set by performing feature selection on all data, then split into a training and test set for model training and evaluation. Finally, we only ``report” accuracies greater than a pre-established threshold (unknown to the recovery algorithm). Our model uses the reported accuracies to estimate the learning curve. This experiment is pivotal as it showcases the model's robustness and validity in replicating real-world ML development scenarios, where overfitting and publication bias are prevalent.

Finally, Experiment 3 uses reported accuracies from published meta-analysis across different digital health applications to estimate the underlying learning curve. Then, for some of the fields where data is available, we identify studies published after the meta-analysis publication date (to ensure they are not included in our algorithm) with relatively large sample-sizes to compare our estimates of accuracy with those published from large data. This experiment allows us to evaluate the model on real data and to estimate the amount of overoptimism across different fields of study.
It is important to note that in this experiment, we extend the concept of learning curves beyond a single model and architecture \cite{ruan2024observational} to encompass a specific field. Our underlying assumption is that the average performance of all ML models within this field can, in expectation, be expressed using the proposed model outlined in Eq. \eqref{eq:parametric_model}.

\subsection*{Experiment 1 (Sampling from the observation model)} 
\label{subsec:syn_data}
In this experiment, we directly sample from the observation model in Eq. \eqref{eq:parametric_model}. We simulate $K = 100$ researcher teams independently developing ML models for a binary classification problem, given a dataset with sample size of $n$. The attained classification accuracies are governed by the parametric model in Eq. \eqref{eq:parametric_model}. We consider the case where there is a pre-defined threshold $\gamma_n$ (unknown to the recovery algorithm) that governs the decision of each of the teams to publish. 

Herein, we highlight the results from two sets of model parameters, with additional cases detailed in the supplementary information. 
\begin{itemize}
    \item Problem 1: $A = 0.78$, $\alpha = -1.24 $, $\beta = -0.76$, $\zeta = 0.45$ and $c_1 = 0.50$.
    \item Problem 2: $A = 0.75$, $\alpha = -0.75 $, $\beta = -0.57$, $\zeta = 0.85$ and $c_1 = 0.40$.
\end{itemize}

We simulate our models using these two sets of parameters to generate 100 sample classification accuracies for various sample sizes $n$, ranging from $20$ to $1000$, accepting only those above the set threshold $\gamma_n$. We aim to recover the true learning curve, only from the reported accuracies.

\begin{figure*}[t]
\centering
\begin{subfigure}[b]{0.48\textwidth}
    \centering
    \includegraphics[width=\linewidth]{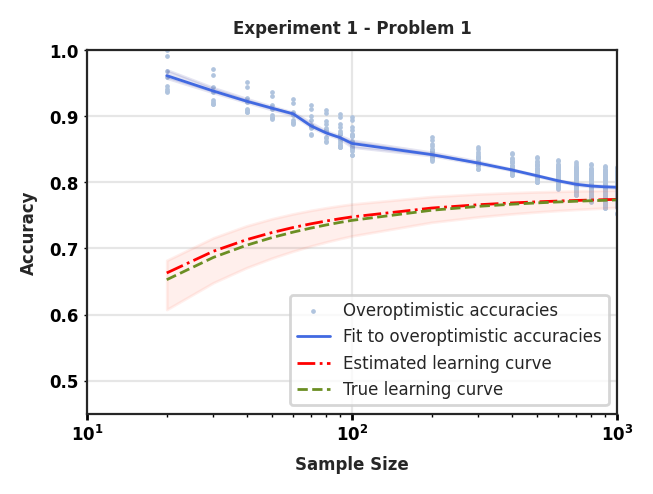}
    \caption{}  \label{subfig:syn_prob1}
\end{subfigure}
\hfill
\begin{subfigure}[b]{0.48\textwidth}
    \centering
    \includegraphics[width=\linewidth]{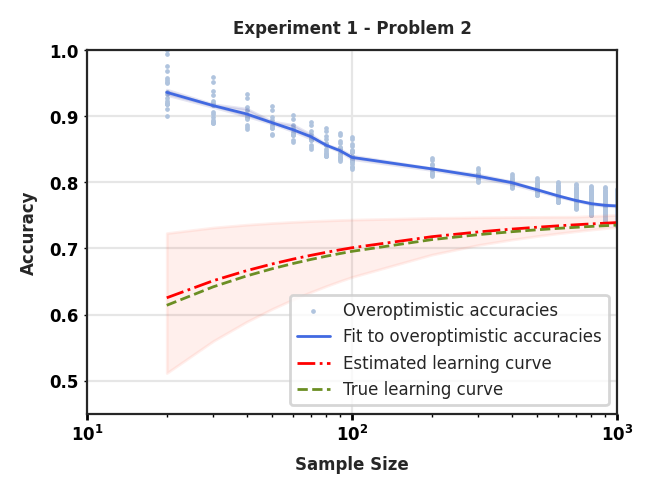}
    \caption{}
    \label{subfig:syn_prob1_sol}
\end{subfigure}
\caption{Evaluation of the proposed method per Experiment 1 where we sample from the observation model for (a) Problem 1: $A = 0.78$, $\alpha = -1.24 $, $\beta = -0.76$, $\zeta = 0.45$ and $c_1 = 0.50$. (b) Problem 2: $A = 0.75$, $\alpha = -0.75 $, $\beta = -0.57$, $\zeta = 0.85$ and $c_1 = 0.40$. The results show the overoptimistic accuracies (blue circles), fit to the overoptimistic results (blue line), the new estimates of the learning curve (red line) along the true learning curve (green line). The y-axis is in linear and x-axis is in log scale.}
\label{fig:synthesized}
\end{figure*}

Fig \ref{subfig:syn_prob1} and \ref{subfig:syn_prob1_sol} show the results, for Problem 1 and Problem 2, respectively. The blue points represent reported accuracies that are inflated due to overfitting and publication bias. The plots further show that the estimated learning curves (red-dashdot lines) and the true learning curves (green-dashed lines) are in close agreement.
\subsection*{Experiment 2 (Simulating overfitting to the test set and publication bias in binary classification)}  We consider two binary classification problems with high dimensional features $\bx$, and  corresponding labels $\bz$. The detailed description of the feature vectors and their corresponding labels are provided in the SI. We consider the scenario where $K = 20$ research teams train ML models, with each team performing feature selection on the complete dataset (prior to train-test split) to simulate a common form of overfitting \cite{vabalas2019machine}. Then, each team randomly selected a training set (70\% of the data) and trained a different ML model. Finally, each team produced an estimate of the accuracy of their classifier on the remaining (30\%) of the data. To account for publication bias, only those accuracies that are greater than a pre-specified threshold $\gamma_n$ are reported. To establish a baseline for comparing the learning curve estimated from only the overoptimistic samples, we attain estimates of the true learning curve by iteratively increasing the sample size and training the model without overfitting to the test set and without publication bias. We repeat this process 100 times and average over these values.

\begin{figure*}[t]
\centering
\begin{subfigure}[b]{0.48\textwidth}
    \centering
    \includegraphics[width=\linewidth]{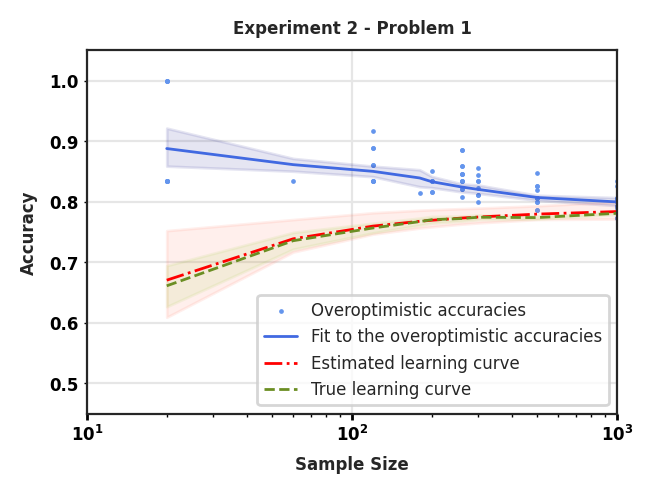}
    \caption{}  \label{subfig:qsyn_prob1}
\end{subfigure}
\hfill
\begin{subfigure}[b]{0.48\textwidth}
    \centering
    \includegraphics[width=\linewidth]{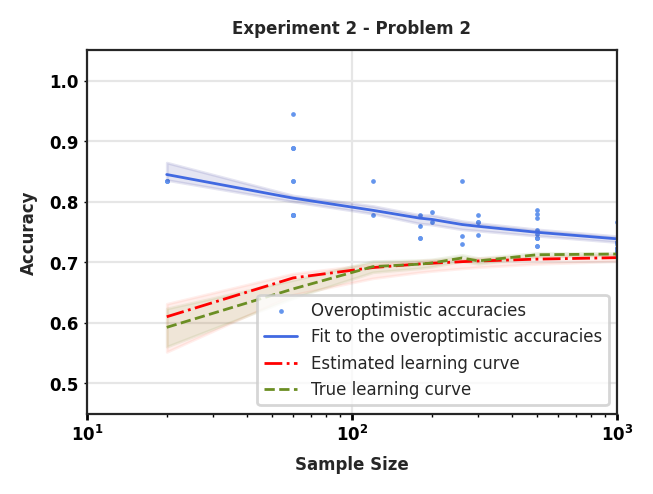}
    \caption{}
    \label{subfig:qsyn_prob2}
\end{subfigure}
\caption{Evaluation of the proposed method per Experiment 2 where we simulate overfitting and publication bias in binary classification, (a) Problem 1, (b) Problem 2. The results show the overoptimistic accuracies (blue circles), fit to the overoptimistic results (blue line), the new estimates of the learning curve (red line) along the true learning curve (green line). The y-axis is in linear and the x-axis is in log scale.}
\label{fig:quasi_syn}
\end{figure*}

Fig. \ref{fig:quasi_syn} shows the overoptimistic classification accuracies (blue circles) and the fit to these published results (solid blue line). The estimated learning curves (red-dashdot lines) is in close agreement with the estimated true learning curves (green-dashed lines). These result provide further backing to our theoretical results in the supplement that the true learning curve is estimable from the overoptimistic estimates using the proposed framework.

\subsection*{Experiment 3 (Estimating the limits of prediction from meta-analyses)} Next, we turn our attention to data from published studies that use ML in a binary classification context. Our aim is to estimate the limits of using ML in a particular field from published results. We focus on published accuracies from meta-analyses of binary classification tasks for prediction of brain disorders \cite{arbabshirani2017single,petti2020systematic,de2020artificial,martinez2021ten,vabalas2019machine}. 
Several studies have reported on the overoptimism in this literature, as evidenced by the negative association in Fig. \ref{fig:negative_asc}. We consider several cases of interest: 
\begin{itemize}
\item Case 1: Classification accuracy of ML models developed using neuroimaging data \cite{arbabshirani2017single} for classifying between patients with Alzheimer's disease (AD) and healthy controls.
\item Case 2: Classification accuracy of ML models developed using speech data \cite{de2020artificial,petti2020systematic,martinez2021ten} for classifying between patients with AD and healthy controls.
\item Case 3: Classification accuracy of ML models developed using neuroimaging data \cite{arbabshirani2017single} for classifying between patients with Autism spectrum disorder (ASD) and healthy controls.
\item Case 4: Classification accuracy of ML models developed using neuroimaging data \cite{arbabshirani2017single} for classifying between patients with attention-deficit hyperactivity (ADHD) and healthy controls.
\item Case 5: Classification accuracy of ML models developed using neuroimaging data \cite{arbabshirani2017single} for classifying between patients with Schizophrenia and healthy controls.
\item Case 6: Classification accuracy of ML models developed using speech data \cite{de2020artificial,petti2020systematic,martinez2021ten} for classifying between patients with forms of cognitive impairment that is not Alzheimer's disease (CI)  and healthy controls.
\item Case 7: Classification accuracy of ML models developed using multi-modality data \cite{vabalas2019machine} for classifying between patients with ASD and healthy controls.
\end{itemize}

\begin{figure*}[t]
\centering
        \begin{subfigure}[b]{0.48\textwidth}
        \centering
        \includegraphics[width=\linewidth]{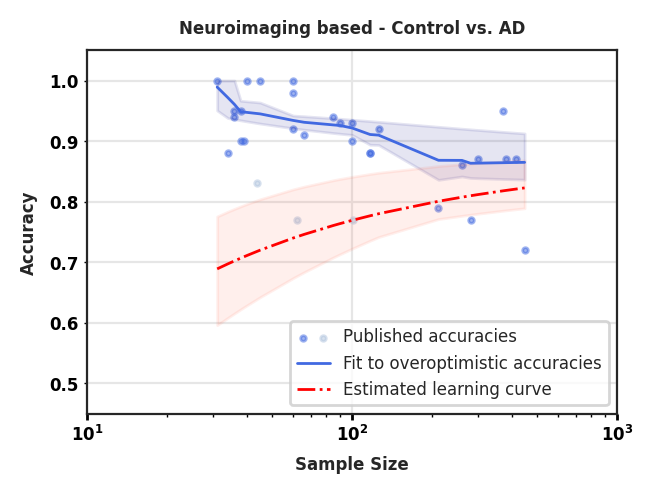}
        \caption{}
        \label{subfig:lc_AD}
    \end{subfigure}
    \hfill
        \begin{subfigure}[b]{0.48\textwidth}
        \centering
        \includegraphics[width=\linewidth]{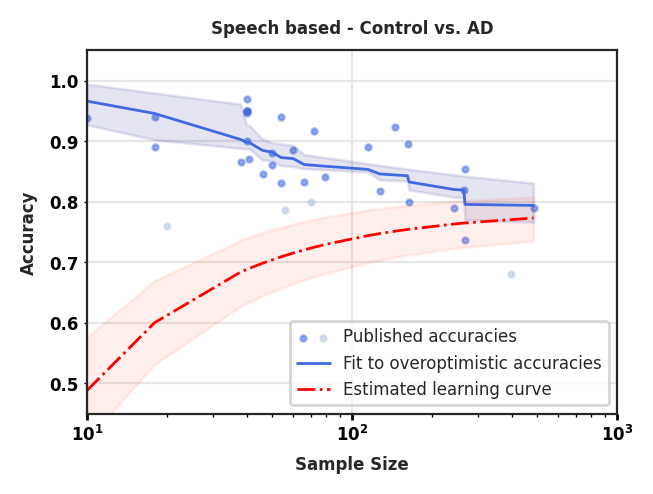}
        \caption{}
        \label{subfig:lc_SP_AD}
    \end{subfigure}
    \caption{Estimates of the learning trajectories of ML models from (a) meta-analyses of neuroimaging-based prediction of AD \cite{arbabshirani2017single} and (b) speech-based prediction of AD \cite{de2020artificial,petti2020systematic,martinez2021ten} per Experiment 3. The results show the new estimates of the empirical learning curve (red), with the published results (blue circles) and fit to the observations (blue). Faded blue circles shown are considered as outliers and were removed from analysis. y-axis is in linear and x-axis is in log scale.}
    \label{fig:real}
\end{figure*}

Here, we present the first two cases and provide the remainder in the SI. Fig. \ref{subfig:lc_AD} and \ref{subfig:lc_SP_AD} illustrates the published classification accuracies (blue circles), and the fit to the overoptimistic results (blue) along with the estimated, de-biased learning curves (red) for the neuroimaging-based and speech based classification of AD, respectively.
As depicted in Fig. \ref{fig:real}, many of the individual studies from the meta-analyses (blue circles) fall outside the upper confidence band of the corrected learning curves (red-dashdot lines); these are identified as overoptimistic published results by the model.

We compare the estimated limiting performance of the parametric model (i.e., parameter $\tilde{A}$) with large-scale studies published after the publication of the meta-analyses above in Table \ref{tab:recent_pub}. In some cases, large-scale studies are not available, as collecting large datasets in these fields can be challenging. Nevertheless, we postulate that the reported performance of ML models trained with larger sample sizes should be more realistic, as the impacts of publication bias and overfitting diminish with increasing sample size. Table \ref{tab:recent_pub} lists the estimated limiting performance across different fields in digital health, including $95 \%$ confidence intervals, alongside the reported accuracy of recent large-scale individual studies with their corresponding sample sizes. The results indicate that reported classification accuracies from larger studies \cite{lu2022practical,yan2019discriminating} fall within the confidence intervals of limiting performance and the estimated learning trajectories for the prediction of AD and Schizophrenia. On the other hand, the reported accuracy for prediction of ASD from \cite{prasad2022deep} is below the anticipated value, estimated by the new, recovered learning curve. This suggests the possibility of further improvement in model development for the large sample-size study published in \cite{prasad2022deep}.

\begin{table*}
\centering
\caption{The estimated limiting performance derived from meta-analyses of the prediction of brain disorders, alongside the published classification accuracy and sample size from individual studies. Abbreviations: HC—Healthy control group, NI—Neuroimaging, AD—Alzheimer's disease, SZ—Schizophrenia, ASD—Autism spectrum disorder}  
\begin{tabular}{lcccccr} \hline
Problem&Limiting Performance&Reference&Sample size&Reported accuracy
\\ \midrule
\multirow{1}{*}{NI-based HC vs. AD}&\multirow{1}{*}{$0.88$, $\left[0.84 - 0.96\right]$}&\cite{lu2022practical}-(2022)&$85721$&$0.91$\\
\multirow{1}{*}{NI-based HC vs. SZ} &\multirow{1}{*}{$0.73$, $\left[0.70 - 0.81\right]$}&\cite{yan2019discriminating}-(2019)&$1100$&$0.80$\\
\multirow{1}{*}{NI-based HC vs. ASD}&\multirow{1}{*}{$0.87$, $\left[0.80 - 0.93\right]$}&
\cite{prasad2022deep}-(2022)&$1026$&$0.75$\\\bottomrule
\end{tabular} 
\label{tab:recent_pub}
\end{table*}

%% file: TEX_files/MaterialsandMethods.tex
\section{Methods}
\label{sec:methods}

In this section, we first discuss the causes behind overoptimism in the ML-based science literature, and the reasoning behind the proposed model. Then, we provide the details of the proposed solution toward alleviating the overoptimism in published results. 
\subsection{Observation model for reported classification accuracy} Learning curves are powerful tools that provide insight into the capabilities of ML models in a specific domain. In the context of supervised learning, the learning curves \cite{mohr2022learning} describe the predictive performance of an ML model by illustrating an estimate of its performance for different sample sizes. One can construct an empirical learning curve by analyzing the classification accuracy of an ML model across a range of sample sizes. A true learning curve accurately represents an ML model's performance for a specific problem, offering insights into how the model's effectiveness changes with the addition of more data samples. Predicting an ML model's capabilities accurately is crucial for guiding future investments and conserving resources by avoiding inefficiencies. However, as demonstrated in Fig. \ref{fig:negative_asc}, the pervasive overoptimism in ML-based science research can obscure the identification of genuine learning curves. This issue of overoptimism, as suggested by the authors in \cite{berisha2021digital}, is likely driven by overfitting and publication bias within the ML scientific community.

\noindent\textit{Overfitting:}
The literature on machine learning (ML) in scientific research highlights the issue of overfitting to the test set, which can result in overly optimistic estimates of classification accuracy. One prevalent practice contributing to this issue is \textit{adaptive data analysis} \cite{dwork2015preserving,dwork2015generalization}, where researchers repeatedly use the same test set to evaluate various models, features, or hyperparameters in an effort to improve model performance.
This approach is widespread in the machine learning community, where maximizing performance metrics are the primary goal. However, this practice becomes problematic in small sample size regimes, as models are more susceptible to learning from noise and unique dataset characteristics, thus increasing the risk of overfitting \cite{roelofs2019meta}. Evidence suggests that reusing a test set multiple times during model development introduces significant overfitting errors, which optimistically scales as $\sqrt{t/ n}$, with $t$ representing the test set's number of reuses \cite{dwork2015preserving,arora2021rip}. Our proposed model, outlined in Eq. \eqref{eq:parametric_model}, integrates this understanding by incorporating Gaussian noise with an elevated mean of $\zeta n^{-0.5}$ to account for overfitting effects.

\noindent \textit{Publication bias:} Inspired by the work on selection bias in econometrics, we model publication bias in similar fashion \cite{amemiya1984tobit}. We assume that $K$ research groups are independently working on a dataset of sample size $n$, with each group building an ML model using this data without influence from other groups, and estimating its accuracy. Let $\by_{n_k}^{\star}$ denote the classification accuracy obtained by the $k$-th research team, where $k = 1,2,\ldots,K$, for datasets of size $n$. Then, we can write:

\begin{equation} \label{eq:tobit_model}
    \begin{aligned}
    \begin{cases}    
 \by_{n_k} = \by_{n_k}^{\star} \text{ for }k = 1,\ldots,K &\text{if} \quad \by_{n_k}^{\star}\geq \gamma_n , \\
    \by_{n_k}^{\star}\text{ not observed  }  \quad &\text{if} \quad \by_{n_k}^{\star} < \gamma_n
    \end{cases}
    \end{aligned} \:,
\end{equation}
where $\by_{n_k}$ denotes the published classification accuracy for team $k$ using sample size $n$ to build and test their model. Eq. \eqref{eq:tobit_model} indicates that accuracy $\by_{n_k}^{\star}$ is observed (published) only if it surpasses a threshold $\gamma_n$; otherwise it remains unobserved (unpublished). Consequently, the count of published accuracies, $M$, is less than the total number of conducted studies and their corresponding accuracies, implying $M\leq K$. Acknowledging that publication bias more significantly affects studies with smaller sample sizes and diminishes as sample size increases, we model $\gamma_n$ as a decreasing function of sample size. In addition, models trained with larger sample sizes generally exhibit higher reliability, and consequently, researchers are more inclined to publish studies with large sample sizes. This further supports the rationale that the threshold $\gamma_n$ should be modeled as a decreasing function of $n$.

The proposed selection model (Eq. \eqref{eq:tobit_model}) highlights publication bias, indicating that some studies' results may remain unpublished. To mitigate this bias and accurately estimate the true learning trajectory of ML models, we employ regression with truncated samples. This approach aims to estimate the statistics of Gaussian distributions from the observed data. We outline a solution for estimating these statistics and the parameters of the learning curve. Additionally, a theoretical analysis on the identifiability of this learning trajectory, through the lens of identifying the mean and variance of truncated Gaussian distributions, is presented in the Supplementary Information (SI).

\subsection{Proposed solution} Assume $x \sim \mathcal{N}\left(\mu,\sigma^2\right)$, then from \cite{cohen1991truncated} we can write
\begin{equation}
\label{eq:mills_ratio}
    \begin{aligned}
        \mathbb{E}\left[x|x>\gamma\right] = \mu + \sigma \frac{\phi\left(\frac{\gamma-\mu}{\sigma}\right)}{1-\Phi\left(\frac{\gamma-\mu}{\sigma}\right)}
    \end{aligned}\:,
\end{equation}
and 
\begin{equation}
\label{eq:var_imr}
    \begin{aligned}
        \mathbf{Var}\left[x|x>\gamma\right]\hspace{-1pt} =\hspace{-1pt} \sigma^2\hspace{-2pt}\left[1 \hspace{-2pt} +  \hspace{-2pt} \frac{\gamma-\mu}{\sigma}\hspace{-1pt} \left[\frac{\phi\left(\frac{\gamma-\mu}{\sigma}\right)}{1\hspace{-1pt}-\hspace{-1pt}\Phi\left(\frac{\gamma-\mu}{\sigma}\right)}\right] \hspace{-3pt} - \hspace{-3pt} \left[\frac{\phi\left(\frac{\gamma-\mu}{\sigma}\right)}{1\hspace{-1pt}-\hspace{-1pt}\Phi\left(\frac{\gamma-\mu}{\sigma}\right)}\right]^2\right]
    \end{aligned}\:,
\end{equation}
where $\phi\left(.\right)$ denotes the probability density function (PDF) of a standard normal distribution, $\Phi\left(.\right)$ is the cumulative density function (CDF) of a standard normal distribution, and $\mathbf{Var}\left[x\right]$ denotes the variance of $x$. The ratio in Eq. \eqref{eq:mills_ratio} is known as the inverse of the Mills ratio \cite{takeshi1985advanced}. Given the parametric model in Eq. \eqref{eq:parametric_model} with $\by^{\star}_n \sim \mathcal{N}\left(A + \alpha n^{\beta}+\zeta n^{-0.5},\sigma_n^2\right)$, from Eq. \eqref{eq:mills_ratio} we can write:
\begin{equation}
\label{eq:mlr_model}
    \begin{aligned}     \mathbb{E}\hspace{-2pt}\left[\by_n|\by_n>\gamma_n\right]\hspace{-2pt} =
\hspace{-2pt} 
        A + \alpha n^{\beta}\hspace{-1pt}+\hspace{-1pt}\zeta n^{-0.5} + \sigma_n\psi\left(n;A,\gamma_n,\alpha,\beta,\zeta,c_1\right)
    \end{aligned}\
\end{equation}
where $\sigma_n = c_1/\sqrt{n}$ and

\begin{equation}
\label{eq:IMR}
    \begin{aligned}       \psi\left(n;A,\gamma_n,\alpha,\beta,\zeta,c_1\right) = 
       \frac{\phi\left(\frac{\gamma_n-A - \alpha n^{\beta}-\zeta n^{-0.5}}{\sigma_n}\right)}{1-\Phi\left(\frac{\gamma_n-A - \alpha n^{\beta}-\zeta n^{-0.5}}{\sigma_n}\right)}
    \end{aligned}\:.
\end{equation}
For brevity, we use the notation $\psi\left(n\right)$ for $\psi\left(n;A,\gamma_n,\alpha,\beta,\zeta,c_1\right)$. The average \textit{published} classification accuracy can be expressed as in Eq. \eqref{eq:mlr_model} where the summation of the first two terms models the true learning curve, i.e., $A+\alpha^{\beta}$, the term $\zeta n ^{-0.5}$ models overfitting, and $\psi\left(n\right)$ in Eq. \eqref{eq:IMR} models the impact of publication bias. From Eq. \eqref{eq:var_imr}, the variance of this estimate is 
\begin{equation}
\label{eq:variance_model}
    \begin{aligned}      \mathbf{Var}\hspace{-2pt}\left[\by_n|\by_n>\gamma_n\right]\hspace{-2pt} = \sigma_n^2\hspace{-3pt}\left(\hspace{-2pt}1 + \hspace{-2pt}\frac{\gamma_n \hspace{-2pt} - \hspace{-2pt}A -\hspace{-1pt} \alpha n^{\beta} \hspace{-1pt}-\hspace{-1pt} \zeta n^{-0.5}}{\sigma_n}\psi\left(n\right)  \hspace{-1pt}- \hspace{-1pt}\psi^2\left(n\right)\hspace{-2pt}\right)
    \end{aligned}.
\end{equation}

Cohen (1991) demonstrated that the method of moments can be used to estimate the parameters of a truncated Gaussian distribution by matching the empirical mean and variance of observed samples to those expected from the truncated distribution \cite{cohen1991truncated}. Building on this concept, we propose the following multi-objective optimization problem that aims to find the best match between the observed data's statistical properties and those of a theoretical truncated Normal distribution:
\begin{equation}
\label{eq:optimization}
    \begin{aligned}  
        &\min_{A,\alpha,\beta,\zeta,c_1}  f_m\left(A,\alpha,\beta,\zeta,c_1\right)\\
    &\text{Subject to}\\ 
    &f_1\left(A,\alpha,\beta,\zeta,c_1\right) \hspace{-2pt}=\hspace{-1pt}\sum_{n\in \mathrm{N}} \hspace{-2pt}\left(\bar{\by}_n \hspace{-1pt}-\hspace{-1pt}A\hspace{-1pt}-\hspace{-1pt}\alpha n^{\beta}\hspace{-1pt}-\hspace{-1pt}\zeta n^{-0.5}\hspace{-1pt}-\hspace{-1pt}\sigma_n \psi\left(n\right)\right)^2\\
    &f_2\left(A,\alpha,\beta,\zeta,c_1\right) \hspace{-1pt} =\\
    &\sum_{n\in \mathrm{N}}\hspace{-2pt}\left(\bar{\sigma}^{2}_n\hspace{-1pt}-\sigma_n^2\hspace{-3pt}\left(\hspace{-2pt}1 + \hspace{-2pt}\frac{\gamma_n \hspace{-2pt} - \hspace{-2pt}A -\hspace{-1pt} \alpha n^{\beta} \hspace{-1pt}-\hspace{-1pt} \zeta n^{-0.5}}{\sigma_n}\psi\left(n\right)  \hspace{-1pt}- \hspace{-1pt}\psi^2\left(n\right)\hspace{-2pt}\right)\hspace{-2pt}\right)^2\\
    &0.5 \leq A \leq 1, -2\leq\alpha\leq -0.5,-1\leq \beta \leq 0\hspace{2pt},0\leq\zeta\leq 1, 0 < c_1 \leq 0.5
    \end{aligned} \:,
\end{equation}
where $\bar{\by}_n$ and $\bar{\sigma}^{2}_n$ are sample mean and sample variance of the reported accuracies given a sample size $n$, and set $\mathrm{N}$ contains all sample sizes used in published studies within a specific field. We use the non-dominated sorting genetic algorithm (NSGAII) \cite{deb2002fast} to solve the non-convex optimization program in Eq. \eqref{eq:optimization}. To find the optimal solution, we first construct the pareto front of non-dominated solutions to this problem. Among this set, we select the optimal solution by using $f_1\left(A,\alpha,\beta,\zeta,c_1\right)$ as the augmented scalarization function (ASF) \cite{wierzbicki1980use}. 

We denote the new estimated learning curve as $\tilde{A} + \tilde{\alpha} n^{\tilde{\beta}}$ where $\tilde{A}$, $\tilde{\alpha}$ and $\tilde{\beta}$ are the estimates of the parameters $A$, $\alpha$ and $\beta$ respectively; we denote the fit to the overoptimistic accuracies using Eq. \eqref{eq:mlr_model}. We build the confidence intervals around these estimates using bootstrapping \cite{efron1986bootstrap,efron1994introduction} where we sampled from the reported accuracies randomly and with replacement to construct a bootstrap sample and repeat this process $10000$ times.

In contrast to traditional regression analyses with truncated samples, which often presuppose knowledge of a threshold $\gamma_n$ \cite{cohen1991truncated,amemiya1984tobit}, our approach does not assume this threshold is known beforehand. We use the maximum likelihood estimate of the threshold, which is the smallest observed value among the data, known as the minimum order statistic \cite{zuehlke2003estimation}. To smooth this statistic, We apply a sliding window technique, with a window length of 2 and a stride of 1, across reported sample sizes, $\gamma_n$, taking the lowest reported accuracy within each window as our threshold.  
\subsection{Limitations}
As demonstrated in the SI, there is no fundamental identifiability issue in estimating the trajectory of machine learning models from overoptimistic observations when many observation are available. However, the theoretical results provide limited insight into the sample size of observations required to recover the parameters. Furthermore, when applying the model to results from meta-analyses, evaluating the resulting estimates is difficult due to the absence of ground truth. Although we recommend comparison with individual, large-scale studies, such studies are often unavailable. Additionally, even when such studies are available, the complete model development process may remain opaque, and it is uncertain whether these models have achieved their maximum performance potential.

%% file: TEX_files/conclusion.tex
\section{Conclusion} \label{sec:conclusion}

There is evidence of a prevalence of overly optimistic results in ML-based science, including in the digital health literature. We posit that this overoptimism stems from publication bias and overfitting, phenomena that distort our understanding of a model's true performance. In this paper, we proposed a novel model based on parametric learning curves to express the reported classification accuracy of ML models, taking into account both overfitting and publication bias. We further proposed a solution based on regression with truncated samples to alleviate the overoptimism in the literature. This novel technique paves the way for a more accurate understanding of the capabilities of ML models in specific fields using  existing, yet overoptimistic, published results. Our results on synthetic data, based on the observation model, demonstrate the success of the proposed solution. Meanwhile, results on real data from meta-analyses in digital health reveal a divergence between the reported accuracies and the newly estimated learning curves. The estimated performance limits and convergence rates of these curves can help reveal the true capabilities of ML models across these fields. Furthermore, they can substantiate trust in published results, especially those obtained with limited samples, should they fall within the confidence intervals of these newly estimated learning curve. 

\section*{Acknowledgments}
This work was supported in part by ONR grant N00014-21-1-2615, the John and Tami Marick Foundation, and the National Science Foundation under awards CCF-2007688 and CCF-2048223.

%% file: TEX_files/Appendix1.tex
In this supplementary information (SI), we begin by presenting a theoretical result concerning the identifiability of the mean and the variance of a truncated Gaussian distribution in the absence of prior knowledge about the truncation set, and we explore the identifiability of the learning trajectory of machine learning (ML) models' performance in the presence of publication bias. Following that, we provide a detailed description of the data used in Experiment 2. Additionally, we present further results on synthetic data generated under the model proposed in Experiment 1. This supplementary results support the effectiveness of our proposed solution in recovering the parameters of the learning curve, assuming that the samples are partially observed due to truncation. Moreover, we include a set of extended results on real published data from various domains in digital health. Table \ref{tab:oi} lists the estimated limiting performance in these
fields, including $95\%$ confidence intervals. We list the studies included in this analysis in Tables \ref{tab:AD}, \ref{tab:SZ}, \ref{tab:ASD}, and \ref{tab:ADHD}. For details on the studies related to speech-based AD and CI, see \cite{berisha2022over}\footnote{http://vees.ar/IS22SpeechMLTable.pdf}. Further information on the studies related to multimodal ASD prediction can be found in the supplementary document cited in \cite{vabalas2019machine}. It should be noted that from the meta-analyses, we included only the published results that address binary classification problems between a healthy control group and a patients group. Studies with multiple groups of patients are excluded from this analysis.
\section{On identifiability of the parameters of a truncated Gaussian distribution}
In this paper, we proposed a parametric model that characterizes the learning trajectory of machine learning (ML) models within a specific domain, reflecting the average performance of ML models across varying sample sizes. We defined this learning trajectory, or learning curve, as $ = A + \alpha n^{\beta}$. There is substantial evidence suggesting that many published results reflect an overestimation of model performance, and we postulated that this phenomenon was primarily due to overfitting and publication bias. Therefore, the published results predominately reveal only the upper end of the performance estimates for these models, thereby failing to represent the true learning trajectory. Therefore, we directed our focus towards regression analysis using truncated samples to accurately identify the true learning trajectory. To that end, we ask: is it possible to accurately estimate this learning curve from the observed samples alone, especially when the truncation thresholds remain unknown? To answer this question, we first properly formulate the problem. 

Let $\mathrm{N}$ denote a set that includes all sample sizes used to train machine learning models within a particular field, with $N = |\mathrm{N}|$ representing the set cardinality. For every $n \in \mathrm{N}$, research groups train and evaluate many machine learning models. The accuracies of these models are assumed to be independently drawn from a Gaussian distribution, $\mathcal{N}\left(\mu_n,\sigma_n^2\right)$, and are subject to a selection mechanism governed by a threshold $\gamma_n$. Consequently, we only observe those samples that exceed the threshold $\gamma_n$. The question we address is whether, for all $n\in\mathrm{N}$, the means, $\mu_n$, and variances, $\sigma_n^2$, of these Gaussian distributions can be identified from the observed samples when these thresholds $\gamma_n$ are unknown. 

Regression with truncated samples is a classical problem in statistics \cite{fisher1931properties,cohen1991truncated} and econometrics \cite{amemiya1984tobit}. The identifiability of the mean and variance in Gaussian distributions has been extensively explored. Notably, the identifiability of the mean vector and covariance matrix in multivariate Gaussian distributions has been established in asymptotic regimes, provided that the truncation set $\mathrm{S}$, which governs the selection mechanism, is known \cite{hotelling1948fitting,tukey1949sufficiency}. Specifically, only those samples that fall within set $\mathrm{S}$ are observed.
Conversely, authors in \cite{daskalakis2018efficient} demonstrated that, in the absence of prior knowledge about the set $\mathrm{S}$, two univariate Normal distributions with distinct means and variances become indistinguishable based solely on the observed samples. However, in our work, we specifically consider cases where the truncation set $\mathrm{S}$ is a half-line space. As we show next, this restriction enables the identifiability of the mean and variance of the Gaussian distribution within finite sample regimes. 

\begin{theorem} \label{th:main}
    Let $\mathrm{N}$ denote the set of all sample sizes used to train ML models in a field. For every $n \in \mathrm{N}$, assume access to $M$ samples (reported accuracies) that are independently drawn from a univariate Gaussian distribution with mean $\mu_n$ and variance $\sigma_n^2$, and have survived a truncation mechanism such that the observations exceed a fixed, yet unknown threshold $\gamma_n$. Provided $M$ satisfies Eq. \eqref{eq:N_finite_bounds} for all $n\in \mathrm{N}$, then, with high probability, it is possible to identify $\left(\tilde{\mu}_n,\Tilde{\sigma}_n^2\right)$ for each $n \in \mathrm{N}$, and consequently, the learning trajectory of the ML models, such that  $\left|\frac{1}{\sigma_n}\left(\mu_n - \Tilde{\mu}_n\right)\right|\leq \epsilon$ and $\left|1 - \frac{\Tilde{\sigma}_n^2}{{\sigma_n^2}}\right|\leq \epsilon$, for any $\epsilon>0$.
\end{theorem}

To prove this theorem we first need the following results.

\begin{lemma}[Special case of Theorem 1 in \cite{kontonis2019efficient}]
\label{th:uni_idf}
Let $\mathrm{S}$ be a family of sets of half-line spaces, that are lower bounded at a fixed but unknown point $\gamma$ and let $\mathcal{N}\left(\mu,\sigma^2,\mathrm{S}\right)$ be a truncated Gaussian distribution such that $\mathcal{N}\left(\mu,\sigma^2;\mathrm{S}\right)\geq \rho$, where $\mathcal{N}\left(\mu,\sigma^2;\mathrm{S}\right)$ denote the total mass that is contained in the set $\mathrm{S}$ by the original distribution $\mathcal{N}\left(\mu,\sigma^2\right)$.  Given $M$ samples with 
\begin{equation}
    \label{eq:N_finite_bounds}
    M = \text{poly}\left(1/\rho\right)\Tilde{O}\left(\frac{1}{\epsilon^2}\right)
\end{equation}
 Then, with probability at least $99\%$, it is possible to identify $\left(\Tilde{\mu},\Tilde{\sigma}^2\right)$ that satisfy $\left|\frac{1}{\sigma}\left(\mu - \Tilde{\mu}\right)\right|\leq \epsilon$ and $\left|1 - \frac{\Tilde{\sigma}^2}{{\sigma^2}}\right|\leq \epsilon$.    
\end{lemma}
\textbf{Proof of Lemma \ref{th:uni_idf}:} 
The above result is a special case of Theorem 1 in \cite{kontonis2019efficient}, which demonstrates that in a multivariate setting and for an unknown truncation set with a finite Vapnik–Chervonenkis (VC) dimension, the mean vector and covariance matrix of the original $d$-dimensional multivariate Gaussian distribution can be identified with high probability. Specifically, the authors of \cite{kontonis2019efficient} established that the total variation distance $d_{TV}$ between the original Gaussian distribution $\mathcal{N}\left(\mathbf{\mu},\mathbf{\Sigma}\right)$ and the estimated distribution $\mathcal{N}\left(\Tilde{\mathbf{\mu}},\Tilde{\mathbf{\Sigma}}\right)$ is less than any arbitrary small positive number $\epsilon$, when the number of observed samples $M$ obeys:

\begin{equation}
    \label{eq:N_finite_bounds_restate}
    M = \text{poly}\left(1/\rho\right)\Tilde{O}\left(\frac{d^2}{\epsilon^2}+\frac{\text{VC}\left(\mathrm{S}\right)}{\epsilon}\right).
\end{equation}

Let $\mathrm{S}$ be a family of half-line spaces, that are lower bounded at a fixed point $\gamma$. Then, this set can be characterized as $\mathrm{S} = \{y|y \geq \gamma\}$, that has $\text{VC}\left(\mathrm{S}\right) = 1$. Then, given a univariate Gaussian distribution with $d = 1$, it directly follows that if $M$ obeys Eq. \eqref{eq:N_finite_bounds},
the mean and variance of a univariate Gaussian distribution are still identifiable with high probability, when only observing the truncated samples and in settings when the truncation point $\gamma$ is unknown.
\qed

\noindent\textbf{Proof of Theorem \ref{th:main}}
The proof follows directly from Lemma \ref{th:uni_idf}.
Let $\epsilon>0$. For every $n \in \mathrm{N}$, consider $M$ observations independently drawn from a truncated Gaussian distribution $\mathcal{N}\left(\mu_n,\sigma_n^2,\mathrm{S}_n\right)$. Assume that the set $\mathrm{S}$ denotes a half-line space characterized by the truncation point $\gamma_n$. According to Lemma \ref{th:uni_idf} if the number of observations $M$ satisfies Eq. \eqref{eq:N_finite_bounds} from Lemma \ref{th:uni_idf}, it is possible to identify both the estimated mean ($\tilde{\mu}_n$) and variance ($\tilde{\sigma}_n^2$) with high probability such that $\left|\frac{1}{\sigma_n}\left(\mu_n - \Tilde{\mu}_n\right)\right|\leq \epsilon$ and $\left|1 - \frac{\Tilde{\sigma}_n^2}{{\sigma_n^2}}\right|\leq \epsilon$.
This result extends to affirm the identifiability of the means and variances of all the Gaussian distributions for every $n \in \mathrm{N}$ with high probability, and consequently, the identifiability of the learning trajectory of ML models.
\qed

It is important to note that the primary objective of this work is to identify the learning trajectory of ML models form the overoptimistic observed samples. The result discussed above demonstrates that publication bias does not represent an identifiability issue in this context. Furthermore, we estimate this learning curve by identifying the parameters of the proposed model. As discussed in the main paper, we proposed that estimates of model accuracy can be expressed, in expectation, as $A + \alpha n^{\beta} + \zeta n^{-0.5}$ in the presence of overfitting. In addition, we incorporated a random variable to account for inherent variability in the ML model development phase, which scales with $c_1 n^{-0.5}$. Thus, the goal is to identify the parameters $A$, $\alpha$, $\beta$, $\zeta$ and $c_1$.
It has been established that if $q_1$ and $q_2$ are two distinct positive proper fractions, then the numbers $a^{q_1}$ and $a^{q_2}$ are linearly independent \cite{besicovitch1940linear}.
From this result, it is straightforward to show that the two exponential terms, $\alpha n^{\beta}$ (with $\beta<0$) and $\zeta n^{-0.5}$ are linearly independent, provided that $\beta$ is bounded away from $0.5$. With this mild assumption, in this work we have formulated an optimization problem based on the method of moments to recover the parameters of the model. Due to the non-convex nature of the objective functions, we used genetic algorithms to solve the optimization problem. 

\section{Data generating process for Experiment 2}\label{subsec:exp2}
In Experiment 2, we consider two binary classification problems with $10$-dimensional feature vectors $\bx$, and corresponding labels $z$. For both problems we train and evaluate a linear classifier. In Problem 1, the feature vectors are randomly and independently drawn from a multivariate Gaussian distribution. For each feature vector $\bx_i = \left[x_1,x_2,\ldots,x_{10}\right]$ we first compute an output variable $y_i$ as
\begin{equation}
    y_i = \frac{1}{1+\exp\left({-\left(0.8x_1 - 0.3x_2+0.7\epsilon\right)}\right)},
\end{equation}
where $\epsilon \sim \mathcal{N}\left(0,1\right)$ and $i = 1,2,\ldots,n$. Then, the corresponding label is as follows:
\begin{equation}
    z_i = \begin{cases}
        1 \quad y_i > 0.5\\
        0 \quad \text{Otherwise}.
    \end{cases}
\end{equation}
Similarly in Problem 2, the feature vectors are randomly and independently drawn from a multivariate Gaussian distribution. However, this time for each $\bx_i$ we first compute an output variable $y_i$ as \cite{friedman1991multivariate,may2008non}

\begin{equation}
    y_i = 5\left(2 \sin \left(\pi x_1x_2\right) + 4\left(x_3-0.5\right)^2+2x_4+x_5\right) + \epsilon.
\end{equation}
Then, the corresponding label is as follows:
\begin{equation}
    z_i = \begin{cases}
        1 \quad \frac{1}{1+\exp\left(y_i - \bar{y}\right)} > 0.5\\
        0 \quad \text{Otherwise}\:,
    \end{cases}
\end{equation}
where $\bar{y}$ represents the average of all $y_i$'s for $i = 1,2,\ldots n$.
\section{Extended results}\label{subsec:extended}
\begin{figure*}

\centering
            \begin{subfigure}[b]{0.48\textwidth}
        \centering
        \includegraphics[width=\linewidth]{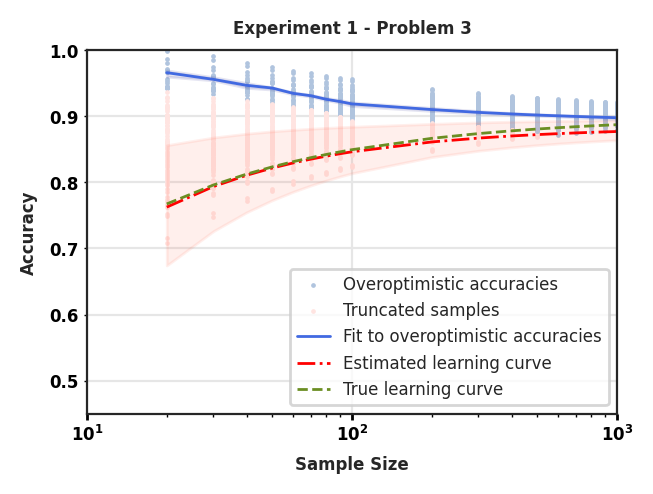}
        \caption{}
        \label{subfig:syn_prob3}
    \end{subfigure}
    \hfill
            \begin{subfigure}[b]{0.48\textwidth}
        \centering
        \includegraphics[width=\linewidth]{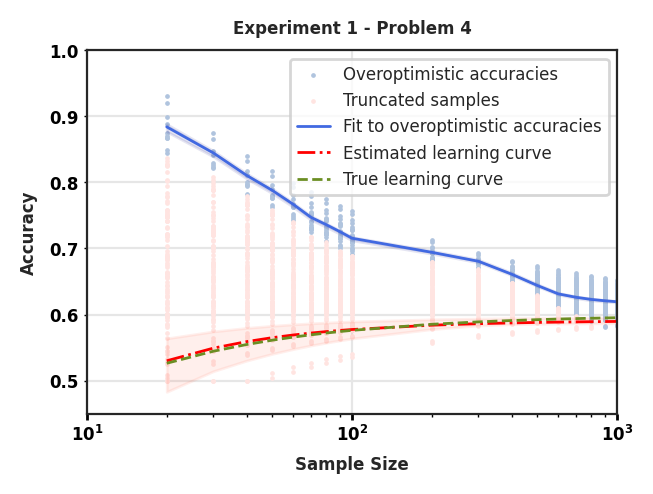}
        \caption{}
        \label{subfig:syn_prob4}
    \end{subfigure}
    \hfill
        \begin{subfigure}[b]{0.48\textwidth}
        \centering
        \includegraphics[width=\linewidth]{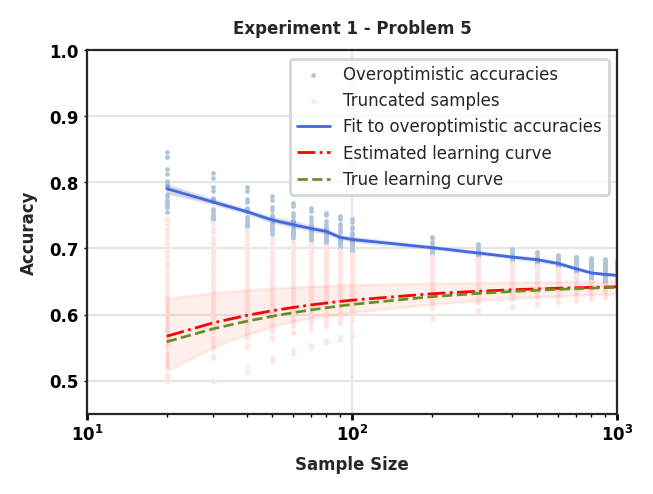}
        \caption{}
        \label{subfig:syn_prob5}
    \end{subfigure}
    \hfill
            \begin{subfigure}[b]{0.48\textwidth}
        \centering
        \includegraphics[width=\linewidth]{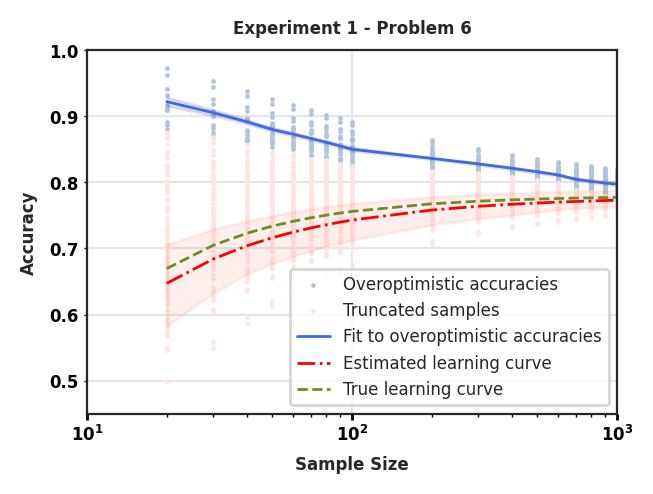}
        \caption{}
        \label{subfig:syn_prob6}
    \end{subfigure}
    \hfill
        \begin{subfigure}[b]{0.48\textwidth}
        \centering
        \includegraphics[width=\linewidth]{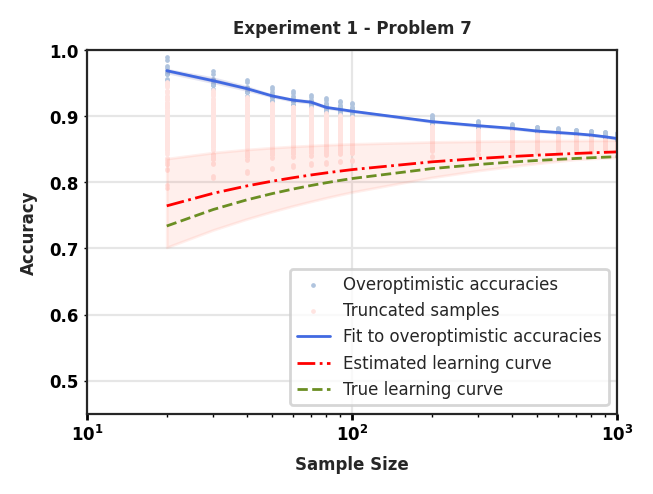}
        \caption{}
        \label{subfig:syn_prob7}
    \end{subfigure}
    \caption{Extended results based on Experiment 1 and for 5 additional cases introduced in Section \ref{subsec:extended}.} 
    \label{fig:extended_synt}
\end{figure*}

We extend our results to additional cases based on Experiment 1, which was introduced in the paper. In this experiment, data is generated based on the proposed model, expressed as 
\begin{equation}
\by^{*}_n = A + \alpha n^{\beta} + \bw_n \:,
\end{equation}
where $\bw_n \sim \mathcal{N}\left(\zeta n^{-0.5},\sigma_n^2\right)$, and $\sigma_n = c_1 n^{-0.5}$, and subsequently subjected to truncation due to the selection mechanism. We categorize each ML problem as `easy' if the performance limit $A > 0.75$, indicating greater separability, and as `difficult' if $A \leq 0.75$. Additionally, the parameter $\beta$ controls the convergence rate, with larger values in absolute value indicating faster convergence. Furthermore, higher values of the overfitting parameter $\zeta$ result in greater overfitting to the test set.
To cover a broad range of scenarios, we consider the following 7 cases:

\begin{itemize}
\item Problem 1 (high separability - fast convergence): $A = 0.78$, $\alpha = -1.24 $, $\beta = -0.76$, $\zeta = 0.45$ and $c_1 = 0.50$.
\item Problem 2 (low separability - slow convergence): $A = 0.75$, $\alpha = -0.75 $, $\beta = -0.57$, $\zeta = 0.85$ and $c_1 = 0.40$.
\item Problem 3 (high separability - slow convergence): $A = 0.90$, $\alpha = -0.80 $, $\beta = -0.60$, $\zeta = 0.40$ and $c_1 = 0.30$.
\item Problem 4 (low separability - fast convergence): $A = 0.60$, $\alpha = -0.60 $, $\beta = -0.70$, $\zeta = 0.70$ and $c_1 = 0.50$.
\item Problem 5 (low separability - slow convergence): $A = 0.65$, $\alpha = -0.55 $, $\beta = -0.60$, $\zeta = 0.40$ and $c_1 = 0.40$.
\item Problem 6  (high separability - fast convergence): $A = 0.78$, $\alpha = -1.90 $, $\beta = -0.95$, $\zeta = 0.25$ and $c_1 = 0.50$.
\item Problem 7  (high separability - slow convergence): $A = 0.85$, $\alpha = -0.70 $, $\beta = -0.60$, $\zeta = 0.70$ and $c_1 = 0.20$.
\end{itemize}
Results corresponding to Cases 1 and 2 are presented in the main paper. Here, we present the results for the additional cases. Fig. \ref{fig:extended_synt} illustrate the result and show observed and truncated samples (blue and light red circles, respectively), fit to the overoptimistic accuracies (blue line), the new estimated learning curves (red-dashdot lines) and the true learning curves (green-dashed lines). As with the results shown in the main paper, there is good agreement between the true and estimated learning curves.

\begin{figure*}

\centering
        \begin{subfigure}[b]{0.48\textwidth}
        \centering
        \includegraphics[width=\linewidth]{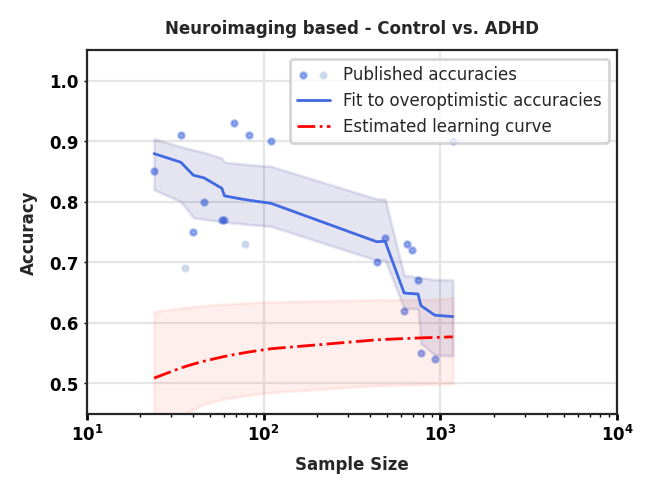}
        \caption{}
        \label{subfig:ADHD}
    \end{subfigure}
    \hfill
        \begin{subfigure}[b]{0.48\textwidth}
        \centering
        \includegraphics[width=\linewidth]{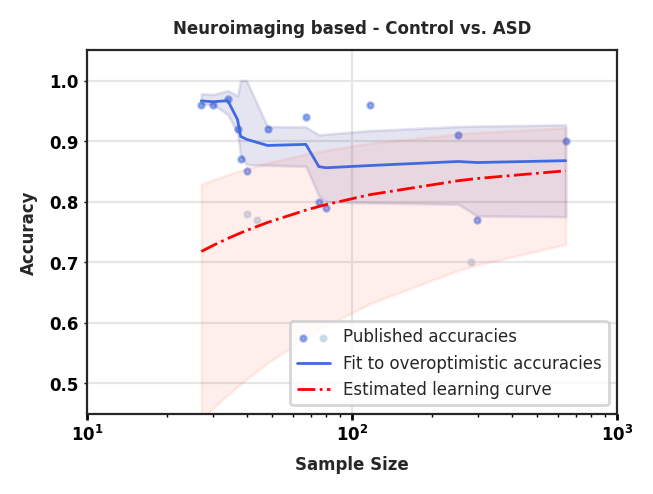}
        \caption{}
        \label{subfig:NIASD}
    \end{subfigure}
    \hfill
        \begin{subfigure}[b]{0.48\textwidth}
        \centering
        \includegraphics[width=\linewidth]{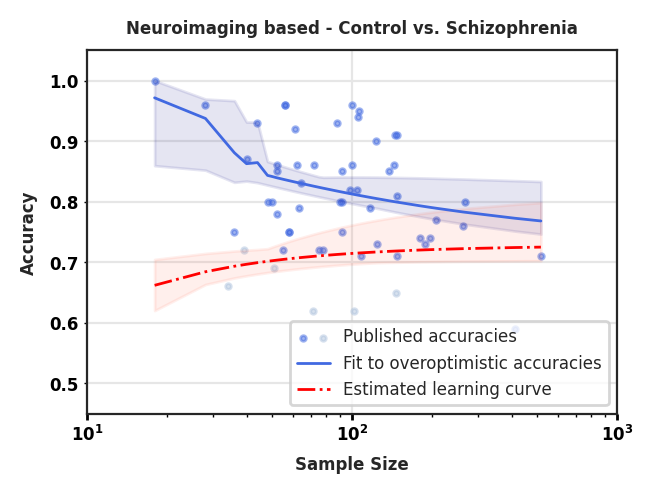}
        \caption{}
        \label{subfig:NISCH}
    \end{subfigure}
    \hfill
        \begin{subfigure}[b]{0.48\textwidth}
        \centering
        \includegraphics[width=\linewidth]{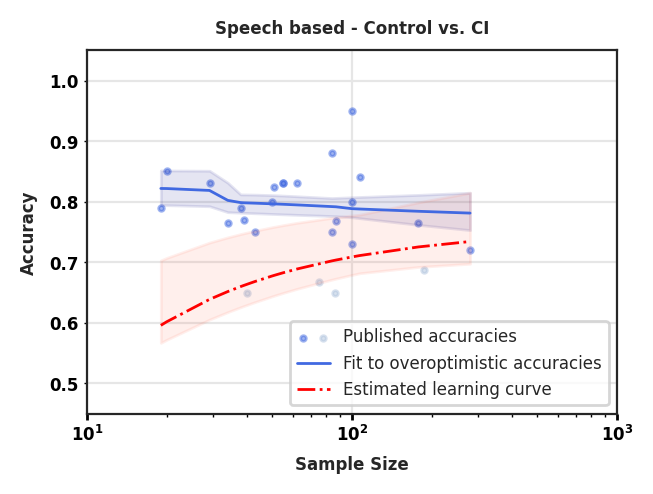}
        \caption{}
        \label{subfig:SPCI}
    \end{subfigure}
    \hfill
        \begin{subfigure}[b]{0.48\textwidth}
        \centering
        \includegraphics[width=\linewidth]{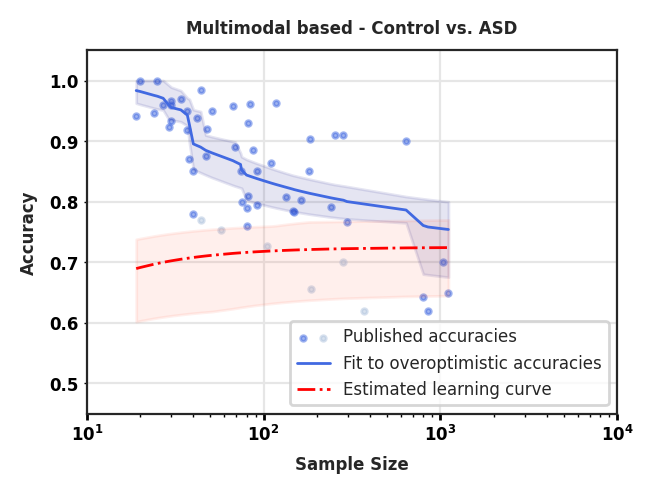}
        \caption{}
        \label{subfig:MMASD}
    \end{subfigure}
    \hfill
    
    \caption{Evaluation of the proposed method on collection of results from meta-analysis of neuroimaging-based prediction of brain disorders \cite{arbabshirani2017single} that includes (a) control vs. ADHD and (b) control vs. ASD, and (c) control vs. Schizophrenia, (d) speech based prediction of CI \cite{petti2020systematic,de2020artificial,martinez2021ten}, and (e) multi-modal based prediction of ASD \cite{vabalas2019machine}. Faded blue are considered as outliers and are removed from analysis. y-axis is in linear and x-axis is in log scale.} 
    \label{fig:lc_all}
\end{figure*}

\begin{figure*}

\centering
        \begin{subfigure}[b]{0.32\textwidth}
        \centering
        \includegraphics[width=\linewidth]{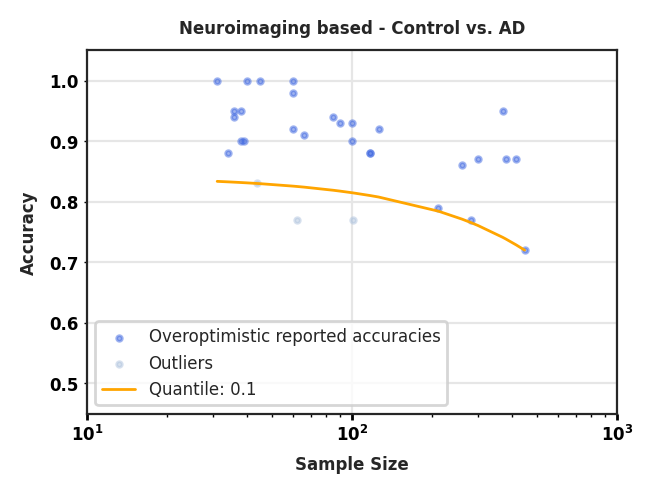}
        \caption{}
        \label{subfig:th_NIAD}
    \end{subfigure}
    \hfill
        \begin{subfigure}[b]{0.32\textwidth}
        \centering
        \includegraphics[width=\linewidth]{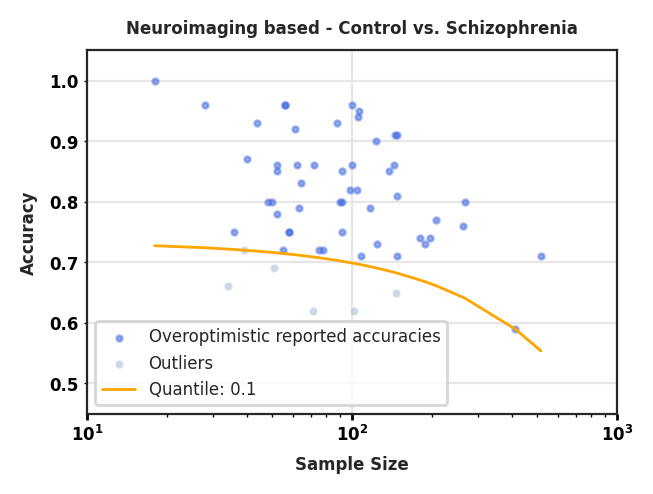}
        \caption{}
        \label{subfig:th_NISZ}
    \end{subfigure}
    \hfill
        \begin{subfigure}[b]{0.32\textwidth}
        \centering
        \includegraphics[width=\linewidth]{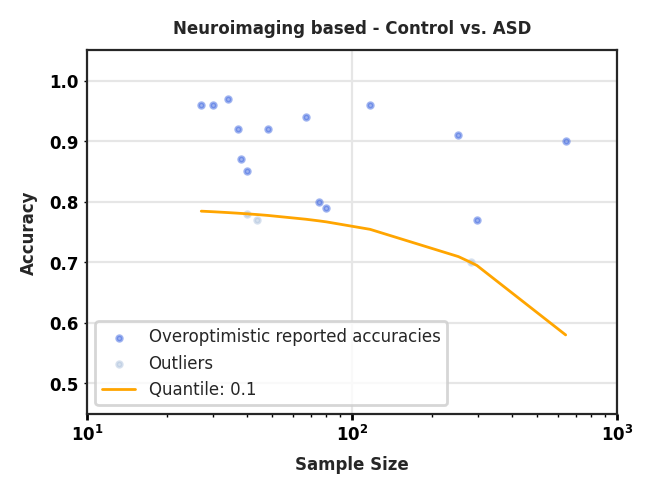}
        \caption{}
        \label{subfig:th_NIASD}
    \end{subfigure}
    \hfill
        \begin{subfigure}[b]{0.32\textwidth}
        \centering
        \includegraphics[width=\linewidth]{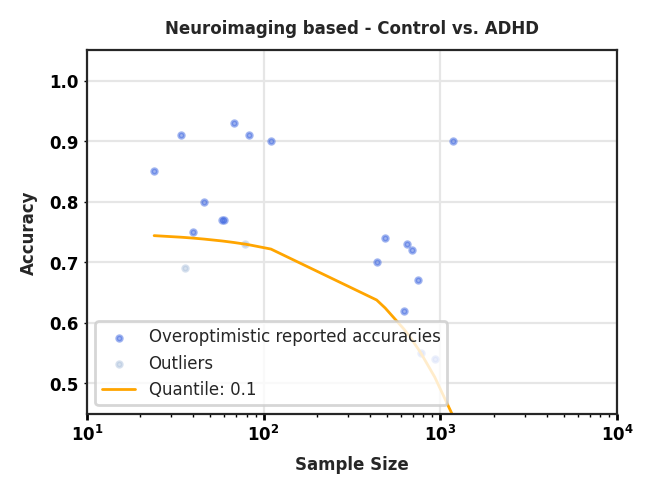}
        \caption{}
        \label{subfig:th_NIADHD}
    \end{subfigure}
    \hfill
        \begin{subfigure}[b]{0.32\textwidth}
        \centering
        \includegraphics[width=\linewidth]{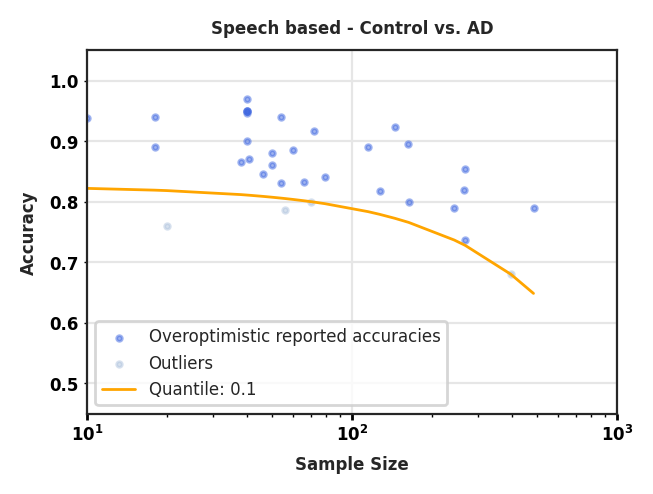}
        \caption{}
        \label{subfig:th_SPAD}
    \end{subfigure}
    \hfill
        \begin{subfigure}[b]{0.32\textwidth}
        \centering
        \includegraphics[width=\linewidth]{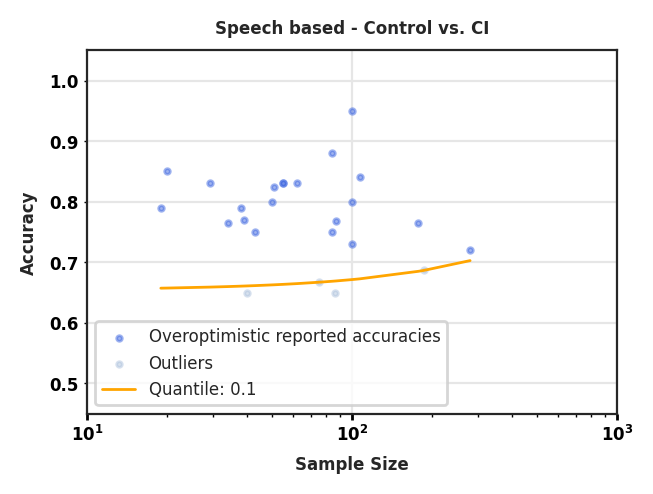}
        \caption{}
        \label{subfig:th_SPCI}
    \end{subfigure}
        \begin{subfigure}[b]{0.32\textwidth}
        \centering
        \includegraphics[width=\linewidth]{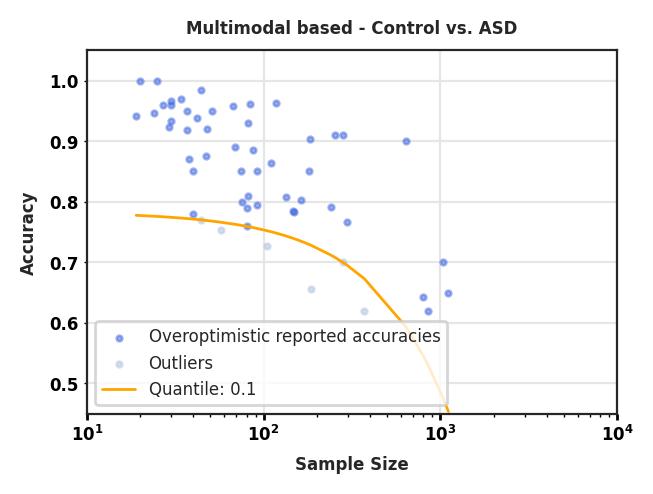}
        \caption{}
        \label{subfig:th_MMASD}
    \end{subfigure}
    \hfill
    \caption{The published results from the meta-analyses \cite{arbabshirani2017single,petti2020systematic,de2020artificial,martinez2021ten,vabalas2019machine} shown with blue circles and address (a) the neuroimaging based control vs. AD, (b) the neuroimaging based control vs. Schizophrenia, (c) the neuroimaging based control vs. ASD, (d) the neuroimaging based control vs. ADHD (e) speech based control vs. AD, (f) speech based control vs. CI and (g) multimodal based control vs. ASD. The orange lines are the $0.1$ quantiles that govern the outlier detection thresholds.} 
    \label{fig:threshold_real}
\end{figure*}

We also provide extended results for Experiment 3. We extend our results to 5 other fields in digital health in addition to neuroimaging and speech based classification of Alzheimer's disease (AD). We consider the following cases:
\begin{itemize}
\item Case 1: Classification accuracy of ML models developed using neuroimaging data \cite{arbabshirani2017single} for classifying between patients with Autism spectrum disorder (ASD) and healthy controls.
\item Case 2: Classification accuracy of ML models developed using neuroimaging data \cite{arbabshirani2017single} for classifying between patients with attention-deficit hyperactivity (ADHD) and healthy controls.
\item Case 3: Classification accuracy of ML models developed using neuroimaging data \cite{arbabshirani2017single} for classifying between patients with Schizophrenia and healthy controls.
\item Case 4: Classification accuracy of ML models developed using speech data \cite{de2020artificial,petti2020systematic,martinez2021ten} for classifying between patients with forms of cognitive impairment that is not Alzheimer's disease (CI) and healthy controls.
\item Case 5: Classification accuracy of ML models developed using multi-modality data \cite{vabalas2019machine} for classifying between patients with ASD and healthy controls.
\end{itemize}
 Fig. \ref{fig:lc_all} shows the published classification accuracies (blue circles), and the fit to the overoptimistic results (blue) along with the estimated, debiased learning curves (red-dashdot lines) across these five fields. The dimmed blue circles in these figures are considered as outliers which have been excluded from our analysis. To identify and remove these outliers, we utilized a quantile regression approach, and discard samples below the $0.1$ quantiles. Fig. \ref{fig:threshold_real} shows the published results and the $0.1$ quantile. 

Lastly, to employ the NSGA II algorithm we used the population size of 40, off springs of 10, the cross-over parameter $\eta_c = 15$ and the mutation parameter $\eta_m = 20$. The total number of generations was varied between 200 to 500 across different fields.

\begin{table}
\centering
\caption{The estimated limiting performance derived from meta-analyses of the prediction of brain disorders \cite{vabalas2019machine,de2020artificial,petti2020systematic,martinez2021ten,arbabshirani2017single}. Abbreviations: HC — Healthy control group, AD — Alzheimer's disease, ASD—Autism spectrum disorder, ADHD—Attention deficit - hyperactivity disorder, CI—forms of cognitive impairment that is not Alzheimer's disease}  
\begin{tabular}{lcccr}
Problem&Limiting Performance\\ \midrule
Neuroimaging-based HC vs. AD& $0.88$, $\left[0.84 - 0.96\right]$
\\
Neuroimaging-based HC vs. Schizophrenia& $0.73$, $\left[0.70 - 0.81\right]$
\\
Neuroimaging-based HC vs. ADHD& $0.59$, $\left[0.50 - 0.68\right]$
\\Neuroimaging-based HC vs. ASD& $0.87$, $\left[0.80 - 0.93\right]$
\\Speech-based HC vs. AD& $0.79$, $\left[0.75 - 0.83\right]$
\\Speech-based HC vs. CI& $0.76$, $\left[0.71 - 0.86\right]$
\\Multi-modal HC vs. ASD& $0.72$, $\left[0.65 - 0.77\right]$
\\
\bottomrule
\end{tabular} 
\label{tab:oi}
\end{table}

\begin{table}
\centering
\caption{List of studies used in Experiment 3 and the analysis for neuroimaging based classification of Alzheimer's Disease \cite{arbabshirani2017single}}  
\begin{tabular}{lcr} 
Sample size& Reported accuracy&Reference\\\midrule
$45$&$1.00$&\cite{grana2011computer}\\
$31$&$1.00$& \cite{wu2012receiver}\\
$40$&$1.00$& \cite{khazaee2015identifying} \\
$100$&$0.90$& \cite{zhang2010optimally} \\
$126$&$0.92$& \cite{zhang2015detection}\\
$44$&$0.83$&\cite{wang2007large}\\
$380$&$0.87$&\cite{vemuri2008alzheimer}\\
$62$&$0.77$&\cite{polat2012computer}\\
$34$&$0.88$&\cite{oliveira2010use}\\
$101$&$0.77$&\cite{miller2009collaborative} \\
$60$&$0.98$& \cite{mahanand2012identification}\\
$38$&$0.95$&\cite{magnin2009support}\\
$39$&$0.90$&\cite{li2007hippocampal}\\
$36$&$0.95$& \cite{lerch2008automated}\\
$117$&$0.88$& \cite{lee2014online}\\
$66$&$0.91$&\cite{freeborough1998mr} \\
$85$&$0.94$&\cite{farhan2014ensemble}\\
$60$&$0.92$&\cite{farzan2015boosting}\\
$370$&$0.95$&\cite{hidalgo2014regions}\\
$60$&$1.00$& \cite{decarli1995discriminant}\\
$100$&$0.93$&\cite{coupe2012simultaneous}\\
$299$&$0.87$&\cite{cuingnet2012spatial}\\
$212$&$0.79$&\cite{chen2014detecting}\\ 
$260$&$0.86$& \cite{beheshti2015probability}\\
$117$&$0.88$& \cite{batmanghelich2011generative}\\
$448$&$0.72$&\cite{adaszewski2013early} \\
$417$&$0.87$&\cite{abdulkadir2011effects}\\
$36$&$0.94$&\cite{li2014discriminative}\\ 

$280$&$0.77$&\cite{dyrba2013robust}\\
$90$&$0.93$&\cite{dukart2013meta}\\
$38$&$0.90$&\cite{dai2012discriminative}\\\bottomrule
\end{tabular} 
\label{tab:AD}
\end{table}

\begin{table}
\centering
\caption{List of studies used in Experiment 3 for neuroimaging based classification of Schizophrenia \cite{arbabshirani2017single}}  
\begin{tabular}{lcr} 
Sample size& Reported accuracy&Reference\\\midrule
$90$&$0.80$&\cite{caprihan2008application}\\
 $58$&$0.75$& \cite{caan2006shaving}\\
$100$&$0.96$& \cite{ardekani2011diffusion} \\
$28$&$0.96$& \cite{honorio2012can} \\
$138$&$0.85$& \cite{demirci2008projection}\\
$34$&$0.66$&\cite{yoon2008multivariate}\\
$102$&$0.62$&\cite{yoon2012automated}\\
$88$&$0.93$&\cite{koch2015diagnostic}\\
$208$&$0.77$&\cite{cao2013integrating}\\
$106$&$0.95$&\cite{castro2011characterization} \\
$52$&$0.85$&\cite{castro2014multiple}\\
$71$&$0.62$&\cite{yu2013functional}\\
$145$&$0.91$&\cite{watanabe2014disease}\\
$92$&$0.75$& \cite{guo2014decreased}\\
$36$&$0.75$& \cite{venkataraman2012whole}\\
$44$&$0.93$&\cite{tang2012identify} \\
$64$&$0.83$&\cite{su2013discriminative}\\
$52$&$0.86$&\cite{shen2010discriminative}\\
$100$&$0.86$&\cite{kim2016deep}\\
$267$&$0.80$& \cite{kaufmann2015disintegration}\\
$48$&$0.80$&\cite{cheng2015nodal}\\
$18$&$1.00$&\cite{fekete2013combining}\\
$62$&$0.86$&\cite{fan2011discriminant}\\ 
$144$&$0.86$& \cite{chyzhyk2015computer}\\
$58$&$0.75$& \cite{bassett2012altered}\\
$56$&$0.96$&\cite{arbabshirani2013classification} \\
$180$&$0.74$&\cite{anticevic2014characterizing}\\
$146$&$0.65$&\cite{anderson2013decreased}\\ 
$148$&$0.71$&\cite{zhang2013optimally}\\
$124$&$0.73$&\cite{zanetti2013neuroanatomical}\\
$92$&$0.80$&\cite{takayanagi2011classification}\\
$72$&$0.86$&\cite{sun2009elucidating}\\
$51$&$0.69$&\cite{radulescu2014grey}\\
$123$&$0.90$&\cite{pina2015predictors}\\ 
$189$&$0.73$& \cite{ota2012discrimination}\\
$516$&$0.71$& \cite{nieuwenhuis2012classification}\\
$104$&$0.82$&\cite{nakamura2004multiple} \\
$262$&$0.76$&\cite{koutsouleris2015individualized}\\
$92$&$0.85$&\cite{kawasaki2007multivariate}\\ 
$78$&$0.72$&\cite{kasparek2011maximum}\\
$75$&$0.72$&\cite{karageorgiou2011neuropsychological}\\
$98$&$0.82$&\cite{janousova2015combining}\\
$39$&$0.72$&\cite{iwabuchi2013clinical}\\
$52$&$0.78$&\cite{ingalhalikar2012identifying}\\
$197$&$0.74$&\cite{greenstein2012using}\\ 
$412$&$0.59$& \cite{gould2014multivariate}\\
$148$&$0.91$& \cite{fan2006compare}\\
$61$&$0.92$&\cite{fan2005classification} \\
$148$&$0.81$&\cite{davatzikos2005whole}\\
$117$&$0.79$&\cite{csernansky2004abnormalities}\\ 
$108$&$0.71$&\cite{castellani2012classification}\\
$105$&$0.94$&\cite{bansal2012anatomical}\\
$50$&$0.80$&\cite{ota2013discrimination}\\
$56$&$0.96$&\cite{du2012high}\\
$55$&$0.72$&\cite{ccetin2015enhanced}\\
$40$&$0.87$&\cite{yang2010hybrid}\\
$63$&$0.79$&\cite{sui2013combination}\\
\bottomrule
\end{tabular} 
\label{tab:SZ}
\end{table}

\begin{table}
\centering
\caption{List of studies used in Experiment 3 and the analysis for neuroimaging based classification of ASD \cite{arbabshirani2017single}}  
\begin{tabular}{lcr}
Sample size& Reported accuracy&Reference\\\midrule
$75$&$0.80$&\cite{ingalhalikar2011diffusion}\\
$34$&$0.97$& \cite{just2014identifying}\\
$27$&$0.96$& \cite{murdaugh2012differential} \\
$40$&$0.78$& \cite{uddin2013salience} \\
$296$&$0.77$& \cite{plitt2015functional}\\
$640$&$0.90$&\cite{iidaka2015resting}\\
$252$&$0.91$&\cite{chen2015diagnostic}\\
$80$&$0.79$&\cite{anderson2011functional}\\
$117$&$0.96$&\cite{wee2014diagnosis}\\
$48$&$0.92$&\cite{uddin2011multivariate} \\
$38$&$0.87$&\cite{jiao2010predictive}\\
$44$&$0.77$&\cite{ecker2010investigating}\\
$40$&$0.85$&\cite{ecker2010describing}\\
$67$&$0.94$& \cite{akshoomoff2004outcome}\\
$30$&$0.96$& \cite{deshpande2013identification}\\
$37$&$0.92$&\cite{libero2015multimodal} \\
$280$&$0.70$&\cite{zhou2014multiparametric}\\\bottomrule
\end{tabular} 
\label{tab:ASD}
\end{table}

\begin{table}
\centering
\caption{List of studies used in Experiment 3 and the analysis for neuroimaging based classification of ADHD \cite{arbabshirani2017single}}  
\begin{tabular}{lcr} 
Sample size& Reported accuracy&Reference\\\midrule
$60$&$0.77$&\cite{hart2014pattern}\\
$34$&$0.91$& \cite{park2016connectivity}\\
$40$&$0.75$& \cite{hart2014predictive} \\
$24$&$0.85$& \cite{zhu2008fisher} \\
$46$&$0.80$& \cite{wang2013altered}
\\
$688$&$0.72$&\cite{sidhu2012kernel}\\
$929$&$0.54$&\cite{sato2012evaluation}\\
$647$&$0.73$&\cite{fair2013distinct}\\
$487$&$0.74$&\cite{dey2012exploiting}\\
$1177$&$0.90$&\cite{deshpande2015fully} \\
$110$&$0.90$&\cite{peng2013extreme}\\
$58$&$0.77$&\cite{lim2013disorder}\\
$68$&$0.93$&\cite{johnston2014brainstem}\\
$78$&$0.73$& \cite{igual2012automatic}\\
$436$&$0.70$& \cite{chang2012adhd}\\
$83$&$0.91$&\cite{bansal2012anatomical} \\
$748$&$0.67$&\cite{anderson2014non}\\
$36$&$0.69$&\cite{iannaccone2015classifying}\\
$624$&$0.62$&\cite{dai2012classification}\\
$776$&$0.55$& \cite{colby2012insights} \\
\bottomrule
\end{tabular} 
\label{tab:ADHD}
\end{table}